\documentclass{llncs}
\usepackage{graphicx}
\usepackage[usenames,dvipsnames,svgnames,table]{xcolor}
\usepackage{hyperref}
\usepackage{bchart} 

\definecolor{linkblue}{RGB}{0,0,155}

\hypersetup{
   bookmarksnumbered,
   colorlinks={true},
   pdfstartview={FitH},
   citecolor={linkblue},
   linkcolor={linkblue},
   urlcolor={linkblue},
}

\newcommand{\pens}[4]{P\textsuperscript{#1}E\textsuperscript{#2}N\textsuperscript{#3}S\textsuperscript{#4}}

\begin{document}

\title{The Controlled Natural Language of\\Randall Munroe's \emph{Thing Explainer}}

\author{
Tobias Kuhn
}

\institute{
  Department of Computer Science, VU University Amsterdam, Netherlands
\smallskip\\
  \texttt{kuhntobias@gmail.com}
}

\maketitle

\begin{abstract}
It is rare that texts or entire books written in a Controlled Natural Language (CNL) become very popular, but exactly this has happened with a book that has been published last year. Randall Munroe's \emph{Thing Explainer} uses only the 1'000 most often used words of the English language together with drawn pictures to explain complicated things such as nuclear reactors, jet engines, the solar system, and dishwashers.
This restricted language is a very interesting new case for the CNL community. I describe here its place in the context of existing approaches on Controlled Natural Languages, and I provide a first analysis from a scientific perspective, covering the word production rules and word distributions.
\end{abstract}

\section{Introduction}

The recent book \emph{Thing Explainer: Complicated Stuff in Simple Words} \cite{munroe2015thingexplainer} by Randall Munroe (who is most well-known for his \emph{xkcd} webcomics) is a very interesting case for the research field of Controlled Natural Languages (CNL) \cite{kuhn2014cl}. It is ``a book of pictures and simple words [...] using only the ten hundred words in our language that people use the most" \cite{munroe2015thingexplainer} and it is both, fun and totally serious. The quote is from the introduction of the book, and therefore it is itself written in this language of only the 1'000 most commonly used English words (and so is, of course, the title of the book). The following paragraph is another example from the section about nuclear power plants, explaining radioactivity:
\begin{quote}
The special heat is made when tiny pieces
of the metal break down. This lets out a lot
of heat, far more than any normal fire could.
But for many kinds of metal, it happens very
slowly. A piece of metal as old as the Earth
might be only half broken down by now. \cite{munroe2015thingexplainer}
\end{quote}
The book has attracted substantial popular interest and press coverage, probably more so than any other book written in a Controlled Natural Language in the recent past, or possibly ever. It has received very positive reviews from prestigious sources such as The New York Times \cite{alter2015nyt}, The Guardian \cite{alderman2015guardian} (``At some points, this produces passages of such startling clarity that one forgets there was ever anything difficult to understand about these phenomena.''), and Bill Gates \cite{gates2015gatesnotes} (``a brilliant concept''), in addition to an interview in New Scientist  \cite{heaven2015newscientist}. But arguably the most flattering review is the one that appeared in The Huffington Post \cite{gleick2015huffingtonpost}, because the journalist himself used the book's controlled language to write the entire review in it! (``So I thought I'd try to tell you a little about this new book the same way, using just those few words.'')
The first edition consisted of 300'000 printed copies \cite{alter2015nyt}, 34'000 of which were sold in the first week alone,\footnote{\url{http://www.publishersweekly.com/pw/by-topic/industry-news/bookselling/article/68882-the-weekly-scorecard-tracking-unit-print-sales-for-week-ending-november-29-2015.html}, retrieved on 9 April 2016} and at the time of writing the book is in the top 20 of best selling books at Amazon in the category \emph{Science \& Math}.\footnote{\url{http://www.amazon.com/Best-Sellers-Books-Science-Math/zgbs/books/75/}, retrieved on 7 April 2016}
This popularity alone makes it an interesting and important CNL to have a closer look at.

The language of \emph{Thing Explainer} is also interesting because of its intriguingly simple restriction applied on top of the English language, namely to use only the top 1'000 most often used words. This kind of simplicity is only rivaled among existing CNLs by the language E-Prime \cite{bourland1965gensem}, whose only restriction is that the verb \emph{to be} is forbidden to use. The fact that the language's restricted vocabulary is not quite as simple as it looks at first, as we will discover below, does not make the concept less intriguing. Most reviewers and readers seem to agree that Randall Munroe succeeds in proving that virtually everything can be explained in an understandable fashion with this so heavily restricted vocabulary.

\section{Language Analysis}

Even though the book and its language have become very popular, not much has been written about the details of the language, the connection to other similar languages, and the precise rules that underlie it. Randall Munroe himself introduced the language in the book with only a few sentences. It is therefore worth taking a closer look here.

\subsection{Similar Languages}

The new language of \emph{Thing Explainer} is similar to some of the earliest English-based CNLs. Basic English \cite{ogden1930basic} was arguably the first such language, presented in 1930. It restricts the allowed words to just 850 root words, but many variations of the language exist. The chosen words and the rules for their application are much more structured than the \emph{Thing Explainer} language, however, allowing for only 18 verbs and imposing substantial restrictions on the grammatical structures within which these words can be used. In this sense, Special English \cite{voa2009wordbook} --- arguably the second oldest English-based CNL --- is more similar. It defines no grammatical restrictions and does not restrict the number of verbs so drastically. It is based on a list of 1'500 words, and has been used since 1959 by the Voice of America, the official external broadcast institution of the United States. In both cases, the words on the list are carefully selected and not just chosen by their frequency in English texts, unlike the \emph{Thing Explainer} language. As another difference, Basic English and Special English define the category for the words on the list, such as \emph{noun} and \emph{verb}, and allow the words only to be used in the given category.
Other similar languages have a more technical background, such as ASD Simplified Technical English (ASD-STE) \cite{asd2013ste}, which also restricts the allowed words and grammatical structures for the aerospace domain, and there are many other similar languages \cite{kuhn2014cl}.

\subsection{Language Properties}

According to the PENS classification scheme, which I proposed in my survey on the topic \cite{kuhn2014cl}, the language of \emph{Thing Explainer} has the same type as Special English, which is \pens1551. This is also the type of full unrestricted English, meaning that the restrictions of the \emph{Thing Explainer} language do not make it considerably different according to the dimensions of the PENS scheme: precision, expressiveness, naturalness, and simplicity.
It is not considerably more precise than full English, because no semantic restrictions come with the restricted vocabulary and the grammar is not restricted at all, and therefore the vagueness and ambiguity of natural language is not significantly mitigated. In terms of expressiveness and naturalness, on the other hand, the power of full English is retained: Randall Munroe comes at least close to proving that basically everything that can be expressed in full English can be expressed in the restricted language as well, in a way that is maybe sometimes funny but always fully natural. With respect to the last dimension, the \emph{Thing Explainer} language is simple when full English can be taken for granted, but it is not significantly simpler than the full language when it has to be described from scratch.
Apart from Special English, other CNLs in the same PENS class are E-Prime, Plain Language, and IBM's EasyEnglish \cite{kuhn2014cl}.

\subsection{Word List}

The list of 1'000 words was assembled by manually merging the word frequency lists from several corpora. Randall Munroe reports it like this:
\begin{quote}
I spent several months going back over a bunch
of different lists and generating some of my
own based on the Google Books corpus and
even my own email inbox. Then I combined
the lists and where they disagreed I just let my
sense of consistency be the tie-breaker. \cite{heaven2015newscientist}
\end{quote}
Conjugated forms of verbs and plural forms of nouns are not listed separately but merged with the plain word form, as Randall Munroe explains: ``I count different word forms---like `talk,' `talking,' and `talked'---as one word.'' This does not apply though to comparative and superlative forms of adjectives (\emph{good}, \emph{better}, and \emph{best} are all separate entries), adverb forms of adjectives (\emph{easy} and \emph{easily} count as two words), or pronouns (\emph{I}, \emph{me}, \emph{my}, and \emph{mine} are separate entries too).
This results in a list of 998 words, which is part of the book, where the missing two words are explained by the fact that ``there's a pair of four-letter words that are very common, but which I left off this page since some people don't like to see them.''

\subsection{Word Production Rules}

Now that we know how the words ended up on the list, let us have a look at how they are supposed to be used from there to write texts such as the book we discuss here. As there are no grammar restrictions, this boils down to selecting word forms from the list and applying word production rules to arrive at related word forms. Randall Munroe's own description of these production rules is very short and not very precise (the first sentence has been quoted above already):
\begin{quote}
In this set, I count different word forms---like ``talk,'' ``talking,'' and ``talked''---as one word. I also allowed most ``thing'' forms of ``doing'' words, like ``talker''---especially if, like ``goer,'' it wasn't a real word but it sounded funny.
\end{quote}
As this description leaves a lot of room for interpretation, we can use the corpus of word forms observed in \emph{Thing Explainer} to reverse engineer the specific rules at work.
Doing so, we arrive at no less than 13 rules, listed here roughly in decreasing order of how naturally they follow from the above description:
\begin{enumerate}
\item \label{rule:list} The word forms on the list of the 1'000 most often used words.
\item \label{rule:v} All conjugation forms of verbs on the list. This includes third singular present (\emph{-s}), past (\emph{-ed}), and infinitive from (\emph{-ing}), including irregular forms.
\item \label{rule:v-er} Noun forms built from verbs on the list by \emph{-er}, for example \emph{carrier}.
\item \label{rule:n-s} The plural forms of nouns on the list (\emph{-s}), for example \emph{things}, including irregular forms like \emph{teeth}. This rule can also be applied to the word \emph{other} to produce \emph{others}, even though it is not a noun.
\item \label{rule:a-er-est} Comparative (\emph{-er}) and superlative forms (\emph{-est}) built from adjectives on the list, for example \emph{smaller} or \emph{fastest}, and including irregular forms like \emph{worse}.
\item \label{rule:n-y-ful} Adjective forms built from nouns on the list by \emph{-y} or \emph{-ful}, for example \emph{pointy} or \emph{colorful}. (The word \emph{colorful} is in fact the only one in \emph{Thing Explainer} derived from this \emph{-ful} production rule.)
\item \label{rule:a-ly} Adverb forms built from adjectives on the list by \emph{-ly}, for example \emph{normally}.
\item \label{rule:a-ness} Noun forms built from adjectives on the list by \emph{-ness}, for example \emph{thickness}.
\item \label{rule:pn} Different case and possessive forms of pronouns on the list: \emph{they} for \emph{them}, \emph{us}/\emph{ours} from \emph{we}/\emph{our}, and \emph{his} from \emph{he}.
\item \label{rule:n-v} Verb forms of nouns on the list and noun forms of verbs on the list when the two forms are similar but not equal, such as \emph{thought} from \emph{think}, and \emph{live} from \emph{life}, including deduced forms like \emph{thoughts} and \emph{living}. (These two cases are in fact the only ones observed for this rule in \emph{Thing Explainer}.)
\item \label{rule:basic} More basic word forms for words on the list, such as nouns from which adjectives on the list were built (\emph{person} from \emph{personal}) and verbs from which nouns on the list were built (\emph{build} from \emph{building}). (Again, these two cases are the only two observed instances of this rule.)
\item \label{rule:a-v} Verb forms built from adjectives on the list, such as \emph{lower} from \emph{low}, including conjugated forms like \emph{lowering} and \emph{lowered}. (This example is again the only observed instance of the rule.)
\item \label{rule:acronym} Common acronyms for words on the list, such as \emph{TV} for \emph{television}. (This example is also the only instance.)
\end{enumerate}
It is not completely clear though, whether some of these later rules point to mistakes rather than features. The confusion of \emph{TV} and \emph{television}, for example, might just be a mistake and not a feature of the language.

\begin{figure}[tb]
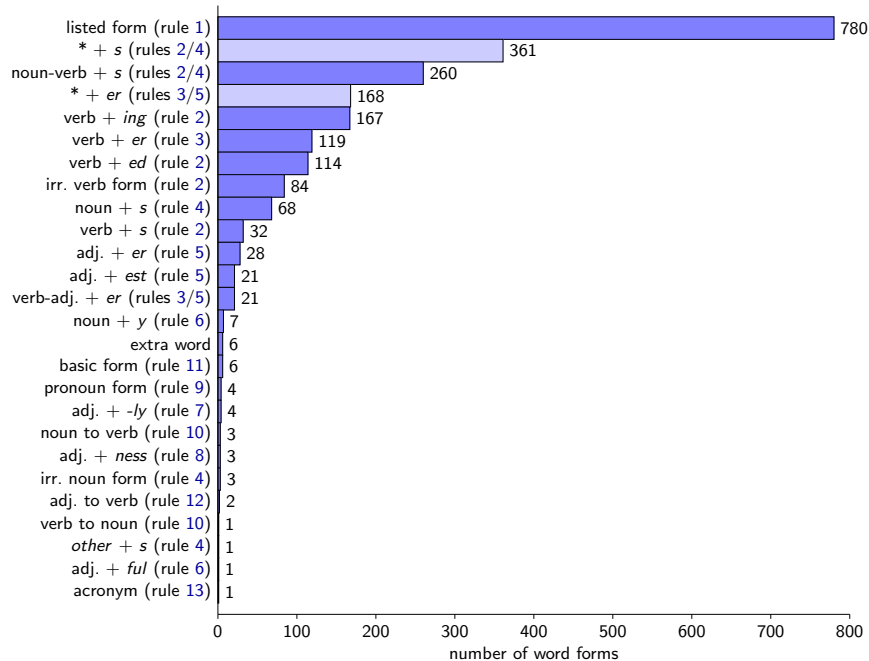

\begin{center}
\scalebox{0.75}{
\begin{bchart}[step=100,max=800,width=14cm,scale=0.8]
  \bcbar[label=listed form (rule \ref{rule:list}),color=blue!50]{780}
  \bcbar[label=* + \emph{s} (rules \ref{rule:v}/\ref{rule:n-s})]{361}
  \bcbar[label=noun-verb + \emph{s} (rules \ref{rule:v}/\ref{rule:n-s}),color=blue!50]{260}
  \bcbar[label=* + \emph{er} (rules \ref{rule:v-er}/\ref{rule:a-er-est})]{168}
  \bcbar[label=verb + \emph{ing} (rule \ref{rule:v}),color=blue!50]{167}
  \bcbar[label=verb + \emph{er} (rule \ref{rule:v-er}),color=blue!50]{119}
  \bcbar[label=verb + \emph{ed} (rule \ref{rule:v}),color=blue!50]{114}
  \bcbar[label=irr. verb form (rule \ref{rule:v}),color=blue!50]{84}
  \bcbar[label=noun + \emph{s} (rule \ref{rule:n-s}),color=blue!50]{68}
  \bcbar[label=verb + \emph{s} (rule \ref{rule:v}),color=blue!50]{32}
  \bcbar[label=adj. + \emph{er} (rule \ref{rule:a-er-est}),color=blue!50]{28}
  \bcbar[label=adj. + \emph{est} (rule \ref{rule:a-er-est}),color=blue!50]{21}
  \bcbar[label=verb-adj. + \emph{er} (rules \ref{rule:v-er}/\ref{rule:a-er-est}),color=blue!50]{21}
  \bcbar[label=noun + \emph{y} (rule \ref{rule:n-y-ful}),color=blue!50]{7}
  \bcbar[label=extra word,color=blue!50]{6}
  \bcbar[label=basic form (rule \ref{rule:basic}),color=blue!50]{6}
  \bcbar[label=pronoun form (rule \ref{rule:pn}),color=blue!50]{4}
  \bcbar[label=adj. + \emph{-ly} (rule \ref{rule:a-ly}),color=blue!50]{4}
  \bcbar[label=noun to verb (rule \ref{rule:n-v}),color=blue!50]{3}
  \bcbar[label=adj. + \emph{ness} (rule \ref{rule:a-ness}),color=blue!50]{3}
  \bcbar[label=irr. noun form (rule \ref{rule:n-s}),color=blue!50]{3}
  \bcbar[label=adj. to verb (rule \ref{rule:a-v}),color=blue!50]{2}
  \bcbar[label=verb to noun (rule \ref{rule:n-v}),color=blue!50]{1}
  \bcbar[label=\emph{other} + \emph{s} (rule \ref{rule:n-s}),color=blue!50]{1}
  \bcbar[label=adj. + \emph{ful} (rule \ref{rule:n-y-ful}),color=blue!50]{1}
  \bcbar[label=acronym (rule \ref{rule:acronym})]{1}
  \bcxlabel{number of word forms}
\end{bchart}}
\end{center}
\caption{The origin of the word forms found in \emph{Thing Explainer} with respect to the different production rules. The total number of such word forms is 1736.}
\label{fig:wordformorigin}
\end{figure}

\begin{figure}[tb]
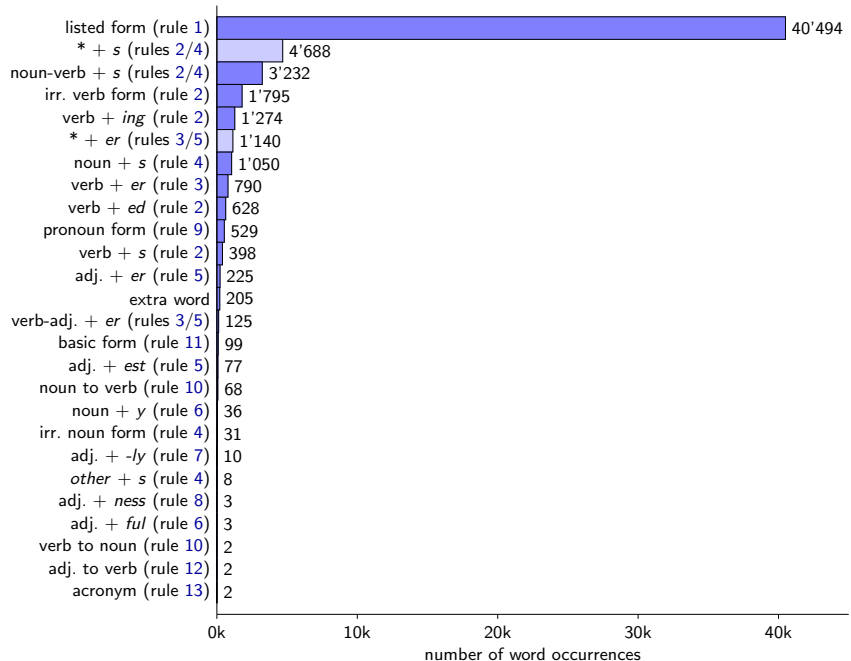

\begin{center}
\scalebox{0.75}{
\begin{bchart}[step=10,max=45,width=14cm,scale=0.8,unit=k]
  \bcbar[label=listed form (rule \ref{rule:list}),color=blue!50,value=40'494]{40.494}
  \bcbar[label=* + \emph{s} (rules \ref{rule:v}/\ref{rule:n-s}),value=4'688]{4.688}
  \bcbar[label=noun-verb + \emph{s} (rules \ref{rule:v}/\ref{rule:n-s}),color=blue!50,value=3'232]{3.232}
  \bcbar[label=irr. verb form (rule \ref{rule:v}),color=blue!50,value=1'795]{1.795}
  \bcbar[label=verb + \emph{ing} (rule \ref{rule:v}),color=blue!50,value=1'274]{1.274}
  \bcbar[label=* + \emph{er} (rules \ref{rule:v-er}/\ref{rule:a-er-est}),value=1'140]{1.140}
  \bcbar[label=noun + \emph{s} (rule \ref{rule:n-s}),color=blue!50,value=1'050]{1.050}
  \bcbar[label=verb + \emph{er} (rule \ref{rule:v-er}),color=blue!50,value=790]{0.790}
  \bcbar[label=verb + \emph{ed} (rule \ref{rule:v}),color=blue!50,value=628]{0.628}
  \bcbar[label=pronoun form (rule \ref{rule:pn}),color=blue!50,value=529]{0.529}
  \bcbar[label=verb + \emph{s} (rule \ref{rule:v}),color=blue!50,value=398]{0.398}
  \bcbar[label=adj. + \emph{er} (rule \ref{rule:a-er-est}),color=blue!50,value=225]{0.225}
  \bcbar[label=extra word,color=blue!50,value=205]{0.205}
  \bcbar[label=verb-adj. + \emph{er} (rules \ref{rule:v-er}/\ref{rule:a-er-est}),color=blue!50,value=125]{0.125}
  \bcbar[label=basic form (rule \ref{rule:basic}),color=blue!50,value=99]{0.099}
  \bcbar[label=adj. + \emph{est} (rule \ref{rule:a-er-est}),color=blue!50,value=77]{0.077}
  \bcbar[label=noun to verb (rule \ref{rule:n-v}),color=blue!50,value=68]{0.068}
  \bcbar[label=noun + \emph{y} (rule \ref{rule:n-y-ful}),color=blue!50,value=36]{0.036}
  \bcbar[label=irr. noun form (rule \ref{rule:n-s}),color=blue!50,value=31]{0.031}
  \bcbar[label=adj. + \emph{-ly} (rule \ref{rule:a-ly}),color=blue!50,value=10]{0.010}
  \bcbar[label=\emph{other} + \emph{s} (rule \ref{rule:n-s}),color=blue!50,value=8]{0.008}
  \bcbar[label=adj. + \emph{ness} (rule \ref{rule:a-ness}),color=blue!50,value=3]{0.003}
  \bcbar[label=adj. + \emph{ful} (rule \ref{rule:n-y-ful}),color=blue!50,value=3]{0.003}
  \bcbar[label=verb to noun (rule \ref{rule:n-v}),color=blue!50,value=2]{0.002}
  \bcbar[label=adj. to verb (rule \ref{rule:a-v}),color=blue!50,value=2]{0.002}
  \bcbar[label=acronym (rule \ref{rule:acronym}),value=2]{0.002}
  \bcxlabel{number of word occurrences}
\end{bchart}}
\end{center}
\caption{The origin of the word occurrences in \emph{Thing Explainer} with respect to the different production rules. The total number of word tokens is 51'086.}
\label{fig:wordorigin}
\end{figure}

Even with these rules, six words remain that are used in \emph{Thing Explainer} but are not allowed according to these rules: \emph{some}, \emph{mad}, \emph{hat}, \emph{apart}, \emph{rid}, and \emph{worth}. It seems that the first one, \emph{some}, should be on the top 1'000 list, but was accidentally omitted. The omission of \emph{they} from the list seems to be a similar mistake. It can be generated by rule \ref{rule:pn} from \emph{them}, but it seems unlikely that \emph{they} would not by itself appear in the top 1'000.
The other five extra words might be explained by the fact that Randall Munroe used a kind of spell-checker while writing to help him use only listed words (``As I wrote, I had tools that would warn me if I used a word that was not on the list, like a spell-checker.'' \cite{heaven2015newscientist}). This spell-checker seems to have over-generated the said words, perhaps because some of them are morphologically close to allowed words: \emph{mad} to \emph{made}; \emph{hat} to \emph{hate}; and \emph{rid} to \emph{ride}.
At other points, Randall Munroe stretches the rules to the extreme. For example, the page about the US Constitution is entitled ``The US's laws of the land'', even though \emph{US} is not on the list, but the pronoun \emph{us} can be inferred from \emph{we} via rule \ref{rule:pn}.

Figures \ref{fig:wordformorigin} and \ref{fig:wordorigin} show the distribution of the word forms and word occurrences, respectively, with respect to the production rules. Only about 45\% of the observed word forms are identical to one on the list (i.e. rule \ref{rule:list}). If individual word occurrences are considered, about 79\% of the 51'086 word tokens in \emph{Thing Explainer} are directly found on the list. This difference is not surprising, considering that the most common words mostly have just one word form (the top 10 most common words in the book are \emph{the}, \emph{to}, \emph{of}, \emph{a}, \emph{it}, \emph{and}, \emph{in}, \emph{that}, \emph{this}, and \emph{you}).
Without parsing the grammatical structure of the sentences, it is not always possible to decide which rule was applied, such as in the case of third singular verb forms being indistinguishable from the plural noun forms for words that could be nouns or verbs. Specifically, 260 word forms ending in \emph{-s} have a root word that could be a noun or a verb (e.g. \emph{names}), and 21 word forms ending in \emph{-er} have a root word that could be a verb or an adjective (e.g. \emph{cooler}).
Apart from the listed word forms (rule \ref{rule:list}), rules \ref{rule:v} to \ref{rule:n-s} are most frequently used.

As a further note, Arabic numerals were not considered for this analysis. It is not clear whether Randall Munroe wanted them to be part of the language or not. He used them only to refer to page numbers, except for two occurrences of \emph{300} in the phrase ``300 years ago'', which he could have written out as ``three hundred,'' as he did in many other places.

\subsection{Word Distribution}

Finally, we can have a look at the word frequency distribution of word forms as they are observed in the book and ``lemmatized'' words as they appear on the list. We can check whether they follow a power law distribution (Zipf's law) as closely as other types of texts \cite{clauset2009siam}. There is no obvious reason a priori why a text in a CNL has to follow the same distribution as an unrestricted one, but it would not be surprising either.

\begin{figure}[tbp]
\begin{center}
\includegraphics[trim=0mm 5mm 0mm 5mm, clip=true, width=0.49\textwidth]{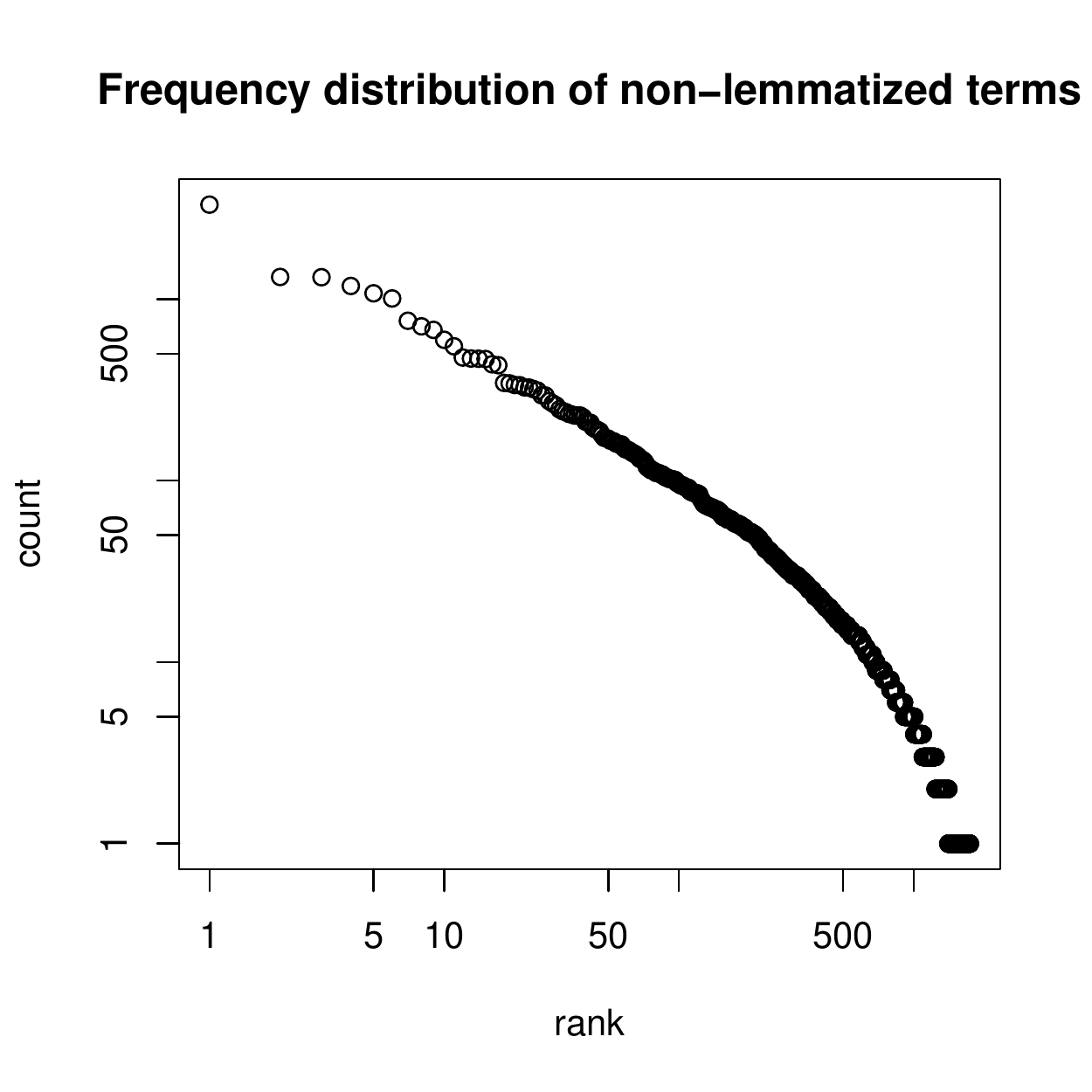}%
\includegraphics[trim=0mm 5mm 0mm 5mm, clip=true, width=0.49\textwidth]{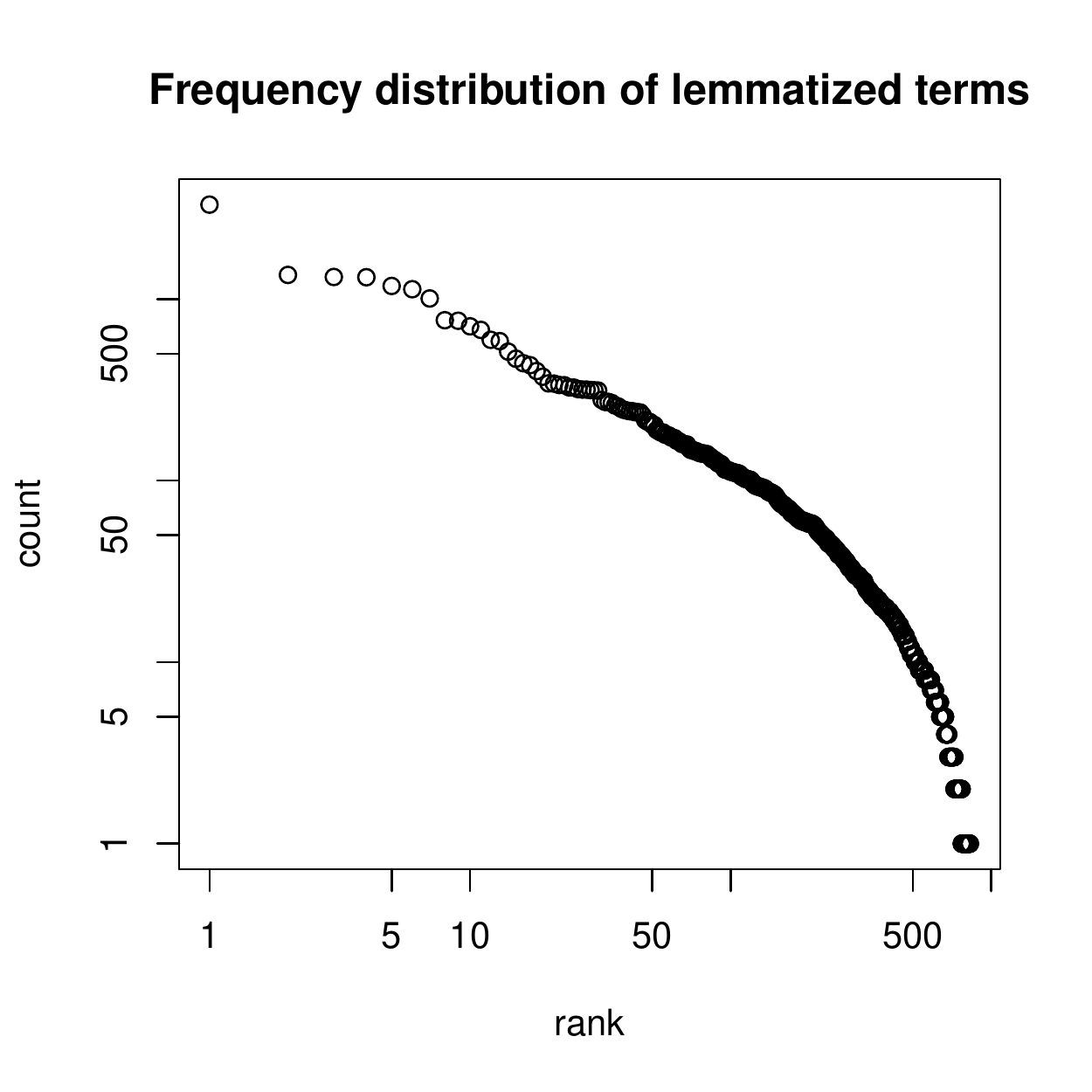}
\end{center}
\caption{The frequency distributions of word forms as they are found in \emph{Thing Explainer} (left), and the distribution of ``lemmatized'' terms when mapped to the list of 1'000 words (right).}
\label{fig:dist}
\end{figure}

The left part of Figure \ref{fig:dist} shows that the distribution of word occurrences follows indeed quite closely a distribution of the kind of a power law, which can be seen by the nearly straight line on the log-log plot. Still the line is curved more than other such word distributions \cite{clauset2009siam}, and therefore tends a bit towards a ``normal'' exponential distribution and away from a pure power law. This effect is even more pronounced in the lemmatized case, as shown on the right hand side part of Figure \ref{fig:dist}. For texts in unrestricted language, it has been shown that the lemmatized distribution is normally very similar to the one of plain word forms \cite{corral2015plosone}.
Still, both distributions are significantly better explained by a power law distribution than an exponential one (with $p$-values of 0.022 and 0.017, respectively). We can hypothesize that such kinds of CNL texts in general mitigate the power law effect as compared to texts in full language, but we cannot make any conclusive statement here.

\section{Methods}

For the analyses presented above, the text is extracted from an electronic version of the book (excluding introduction pages and word list). The book also contains hand written parts, which are not covered. Then all characters except letters, hyphens, and apostrophes are dropped; letters are transformed to lower case; some text extraction errors in the form of missing and extra blank spaces are fixed; words are de-hyphenated; contracted words like \emph{don't} are expanded; \emph{an} is normalized to \emph{a}; Saxon genitive markers \emph{'s} are dropped; the text is tokenized at white spaces; and compound words are split (if the compounds are recognized words). The resulting list of tokens then serves as the input for the analyses.
Furthermore, WordNet is used to detect the categories of words and to lemmatize irregular forms. For the power law analysis, the Python package \texttt{powerlaw} \cite{alstott2014plosone} is used.

\section{Discussion}

While the topics covered by \emph{Thing Explainer} are entirely serious and the book attempts and (I think) succeeds in seriously conveying complex topics in a highly understandable fashion, the book is also fun and the result of a challenge of the sort \emph{how far can we go}.
Randall Munroe admits in the book that ``In some places, I didn't use words even when they were allowed. I could have said `ship,' but I stuck to `boat' because `space boat' makes me laugh.'' At another place, he writes ``light drink that wakes you up'' and ``dark drink that wakes you up,'' even though both, \emph{tea} and \emph{coffee}, are on the list.
On the other hand, there are a number of simple and important words that are not on the list. Randall Munroe reports: ``I could have made it easier for myself. There are a few words I was disappointed didn't make the cut. The biggest omission was a synonym for `rope' or `string'. [...] the only word I had was `line'. This fits in some contexts, but has so many other meanings that it was hard to work with.'' \cite{heaven2015newscientist}
Other examples where the omission of a word rather leads to confusion than simplification include ``the one after eight'' for \emph{nine}, ``white stuff, like what we put on food to make it better'' for \emph{salt}, and ``dirt branch'' for \emph{root}, apart from the omission of proper names to refer to things like countries or planets.
This seems to point to a general problem of such languages with a heavily restricted vocabulary: Writers are forced to circumscribe existing concepts instead of naming them or to involve rough analogies, which can lead to a language that even less precise than full natural language.

These deficiencies could be accounted for to make the language even more useful --- at the expense of some of the funniness --- by increasing the number of words from 1'000 to, perhaps, 1'500 (like Special English) or even 2'000 or 3'000, or by selecting the words manually (again like Special English) instead of being mainly led by their frequency.
There is also a slight inconsistency with respect to how the list of 1'000 words is generated and how it is used. Comparative and superlative forms of adjectives count as separate words when the list is defined, but then these forms can be used in the text even if only the plain form appears on the list. The same applies to adverb forms of adjectives built by \emph{-ly} and pronouns. Normalizing them as well when the list is generated would free some slots for additional words within the limit of 1'000 words.

In general, however, \emph{Thing Explainer} and its language seem to be a huge success, and this success might yield momentum to the general concept of Controlled Natural Language and existing approaches in this field.

\bibliographystyle{abbrv}
\bibliography{thingexplainercnl}

\end{document}